\documentclass[sigconf]{acmart}

\usepackage{amsthm}
\newtheorem{problem}{Problem}
\newtheorem{definition}{Definition}[section]
\newtheorem{theorem}{Theorem}[section]
\newtheorem{proposition}[theorem]{Proposition}

\usepackage{float}
\newenvironment{myitemize}{\begin{list}{$\bullet$}
{\setlength{\topsep}{1mm}
\setlength{\itemsep}{0.25mm}
\setlength{\parsep}{0.25mm}
\setlength{\itemindent}{0mm}
\setlength{\partopsep}{0mm}
\setlength{\labelwidth}{15mm}
\setlength{\leftmargin}{4mm}}}{\end{list}}
\usepackage[ruled,linesnumbered]{algorithm2e}
\SetKwRepeat{Do}{do}{while}

\AtBeginDocument{%
  \providecommand\BibTeX{{%
    \normalfont B\kern-0.5em{\scshape i\kern-0.25em b}\kern-0.8em\TeX}}}

\setcopyright{acmcopyright}
\copyrightyear{2022}
\acmYear{2022}
\acmConference[Milan '22]{Milan '22: ACM/IEEE International Conference on Cyber-Physical Systems}{May 04--06, 2022}{Milan, Italy}
\acmBooktitle{Milan '22: ACM/IEEE International Conference on Cyber-Physical Systems, May 04--06, 2022, Milan, Italy}
\acmPrice{15.00}
\acmISBN{978-1-4503-XXXX-X/18/06}

\begin{document}

\title[Physics-Aware Safety-Assured Design of Hierarchical Neural Network based Planner]{Physics-Aware Safety-Assured Design of \\Hierarchical Neural Network based Planner}

\author{Xiangguo Liu}
\orcid{0000-0002-1944-3923}
%\email{webmaster@marysville-ohio.com}
\affiliation{%
  \institution{Northwestern University}
  %\streetaddress{P.O. Box 1212}
  \city{Evanston}
  \state{IL}
  \country{USA}
  \postcode{60201}
}
\email{xg.liu@u.northwestern.edu}

\author{Chao Huang}
\affiliation{%
  \institution{University of Liverpool}
  %\streetaddress{1 Th{\o}rv{\"a}ld Circle}
  \city{Liverpool}
  \country{UK}}
\email{chao.huang2@liverpool.ac.uk}

\author{Yixuan Wang}
\affiliation{%
  \institution{Northwestern University}
  \city{Evanston}
  \state{IL}
  \country{USA}
}

\author{Bowen Zheng}
\affiliation{%
 \institution{Pony.ai}
 %\streetaddress{Rono-Hills}
 \city{Fremont}
 \state{CA}
 \country{USA}}

\author{Qi Zhu}
\affiliation{%
  \institution{Northwestern University}
  \city{Evanston}
  \state{IL}
  \country{USA}
}

\renewcommand{\shortauthors}{Liu, et al.}

\begin{abstract}
Neural networks have shown great promises in planning, control, and general decision making for learning-enabled cyber-physical systems (LE-CPSs), especially in improving performance under complex scenarios. However, it is very challenging to formally analyze the behavior of neural network based planners for ensuring system safety, which significantly impedes their applications in safety-critical domains such as autonomous driving.  
%However, the main challenge is such neural network based planners are hard to ensure system safety, especially for those safety-critical and near-accident scenarios, such as unprotected left turn. 
%In those scenarios, dramatically different behaviors should be performed to avoid accidents even if under small disturbance, which is even hard for human to ensure safety.
In this work, we propose a hierarchical neural network based planner that analyzes the underlying physical scenarios of the system and learns a system-level behavior planning scheme with multiple scenario-specific motion-planning strategies. We then develop an efficient verification method that incorporates overapproximation of the system state reachable set and novel partition and union techniques for formally ensuring system safety under our physics-aware planner.
%method for scenario-specific formal safety verification based on overapproximation of system state reachable set. 
%motivated by the physical property of such system and develop a method to overapproximate the reachable set to ensure system safety. 
With theoretical analysis, we show that considering the different physical scenarios and building a hierarchical planner based on such analysis may improve system safety and verifiability. We also empirically demonstrate the effectiveness of our approach and its advantage over other baselines in practical case studies of unprotected left turn and highway merging, two common challenging safety-critical tasks in autonomous driving. 
%single neural network planner either makes the system unsafe, or extremely difficult to verify. 
%Thus, we first propose a hierarchical planner composed of behavior planner and multiple motion planners, given that this kind of system may evolve to disjoint subsets of system states and require dramatically different behaviors. 
%The most appropriate motion planner will be selected by behavior planner and thus control the system trajectory to avoid unsafe set. Then we develop the partition and union algorithm to improve scalability and accuracy for reachability analysis and verification.
%For our proposed hierarchical planner, we develop the method to derive overapproximation of reachable set of the system and prove it is sound. 
%Finally we take the unprotected left turn for case study, the simulation and verification results are consistent with our theoritical analysis. The system with our proposed hierarchical planner is safety-assured, while the system with single neural network based planner is unsafe.
\end{abstract}

\begin{CCSXML}
<ccs2012>
   <concept>
       <concept_id>10010520.10010553.10010554.10010557</concept_id>
       <concept_desc>Computer systems organization~Robotic autonomy</concept_desc>
       <concept_significance>500</concept_significance>
       </concept>
   <concept>
       <concept_id>10011007.10011074.10011099.10011692</concept_id>
       <concept_desc>Software and its engineering~Formal software verification</concept_desc>
       <concept_significance>500</concept_significance>
       </concept>
 </ccs2012>
\end{CCSXML}

\ccsdesc[500]{Computer systems organization~Robotic autonomy}
\ccsdesc[500]{Software and its engineering~Formal software verification}

\keywords{physics-aware, safety-assured, neural network, hierarchical planner}

\maketitle

\section{Introduction}
Neural network based machine learning techniques have been increasingly leveraged in learning-enabled cyber-physical systems (LE-CPSs)~\cite{tuncali2018reasoning,wang2020energy} for perception, prediction, planning, control, etc. In particular, neural networks may greatly improve performance and efficiency for planning and general decision making in LE-CPSs, such as autonomous driving~\cite{ha2020leveraging}, human robot interaction~\cite{fisac2018probabilistically}, smart grid~\cite{lu2019incentive} and smart buildings~\cite{wei2017deep}. Moreover, compared with traditional model-based approaches, they can save the time and effort of explicitly modeling systems with complex dynamics and significant uncertainties. However, a major challenge for the neural network based planners is to ensure system safety, especially for safety-critical applications~\cite{zhu2021safety} and in near-accident scenarios~\cite{fan2020bayesian,cao2020reinforcement}, such as unprotected left turn and highway merging in autonomous driving. In those scenarios, with only minor changes in environment states, dramatically different behaviors may need to be performed to avoid accidents, which is difficult for both humans and autonomous systems to handle. Our work focuses on addressing such complex scenarios in safety-critical systems.

In this work, we first observe that for systems that may evolve into different physical scenarios\footnote{For instance, in unprotected left turn, a vehicle may make the turn, yield or stop based on the situation.} under a single neural network based planner, it is often difficult to verify their safety or the planner is indeed unsafe. And we conduct theoretical analysis to show the reason.  
%conduct theoretical analysis for those safety-critical systems under a single neural network based planner. We show that it is either unsafe, or extremely difficult to verify its safety. 
Based on such observation and the fact that many safety-critical systems may evolve into multiple different physical scenarios and thus require dramatically different behaviors to ensure their safety and improve efficiency, we propose a \emph{hierarchical neural network based planner} that consists of a system-level behavior planner and multiple scenario-specific motion planners. We then develop an \emph{efficient verification method} that incorporates novel partition and union techniques and an approach for overapproximating system state reachable set to formally verify the system safety under our hierarchical planner. 
More specifically, our planner design and verification method address the key open challenges in ensuring the safety of neural network based planners, as follows.
%\begin{myitemize}
%\item 

\smallskip
\noindent
\textbf{Hierarchical Planner Design:} We propose a hierarchical planner design for safety-critical systems to improve both system safety and verifiability. Recently, a variety of neural network based planner designs, including hierarchical planners, have been developed for various applications due to their strengths in improving system performance and reducing accident rate in average~\cite{stock2015online,cao2020reinforcement,pettet2020hierarchical}. However, it is very challenging to  formally provide safety guarantee in the worst case for those systems. In particular, we consider the safety-critical systems that may evolve into different physical scenarios based on the dynamic situation, which under a well-designed planner should lead into multiple disjoint subsets of the system state reachable set. We leverage the concept of Lipschitz constant in our theoretical analysis and show that for such systems, most single neural network based planners are either unsafe or will result in significantly harder verification problems when the Lipschitz constant goes to infinity. This motivates our design of a hierarchical planner, which 
consists of a system-level behavior planner and multiple low-level motion planners, each of which corresponds to an underlying physical scenario for the system. %Different from other designs, the number of motion planners $N$ depends on the number of underlying physical scenarios of the system. 
As shown later, our hierarchical planner design, combined with the corresponding improvement in the verification tool, enables formal verification and assurance of system safety.

%We can formally verify the safety of our hierarchical neural network based planner, which cannot be achieved using a single neural network based planner.
%In our approach, we first analyze that those safety-critical systems may evolve into multiple different scenarios, and thus leading to multiple disjoint subsets of system state reachable set. We then leverage the concept of Lipschitz constant to reach the conclusion that most single neural network based planners are indeed unsafe, and it leads to significantly harder verification problems when Lipschitz constant goes to infinity. On the contrary, our hierarchical planner may overcome these difficulties and assure system safety.
%\item 

\smallskip
\noindent
\textbf{Efficient Verification:} Our design of the hierarchical neural network based planner 
brings significant challenges but also opportunities for system safety verification.
%(compared with verifying single neural network based planners). 
In particular, reachability analysis is a popular formal technique for verifying system safety, with various recent methods for LE-CPSs~\cite{tran2019safety,huang2019reachnn,dutta2019reachability,ivanov2019verisig}. However, these methods cannot be directly applied to systems under our hierarchical planner, and have limited efficiency and accuracy for safety-critical systems that are sensitive to environment changes. %, which often lead to disjoint subsets of reachable set, which do not necessarily converge. 
To address these challenges, we first develop new partition and union techniques to overcome the limitations in efficiency and accuracy, and then develop a verification method based on the overapproximated reachable set of both behavior planner and selected motion planners for ensuring system safety.
%\end{myitemize}

\smallskip
\noindent
\textbf{Related Work:}
Our work is related to a rich literature of planner design and system safety verification.
There are a number of varied planner designs, including classical rule-based~\cite{farag2019complex}, optimization-based~\cite{liu2020trajectory} and game theory-based~\cite{liu2020impact} planners, as well as emerging neural network based planners. Many recent neural network based planners demonstrate significant performance improvement and accident rate reduction in average over traditional model-based methods. Some of those learn a single neural network for planning via reinforcement learning~\cite{cao2020highway}, imitation learning~\cite{chen2019deep}, supervised learning~\cite{markolf2020trajectory}, etc., while others employ a hierarchical planner design~\cite{zheng2017generating,cao2020reinforcement}, which usually consists of low-level planners for different modes and a high-level planner that is responsible for selecting the mode. However, even though safety improvement is often considered and demonstrated empirically through experiments in those works~\cite{naveed2020trajectory,cao2020reinforcement,nosrati2018towards,li2021safe,wang2020learning,ma2021model}, formal system safety verification remains a challenging problem. In contrast, our work focuses on formal safety verification, with a hierarchical neural network based planner design that considers the different underlying physical scenarios a system may evolve into. 
%In contrast, our work formally shows that in principle, hierarchical planners are superior over single neural network based planners with respect to safety and verifiability, for safety-critical systems that may evolve to different scenarios.

In terms of safety verification techniques, most recent works present results in ensuring safety for relatively simple scenarios, such as adaptive cruise control and emergency braking~\cite{tran2019safety,huang2020opportunistic}. Different from these works, we focus on safety verification for complex systems that may evolve into multiple different physical scenarios and thus have multiple disjoint reachable set. Our approach verifies systems by computing a bounded time reachable set. However, different from the verification of neural network controlled systems in the literature~\cite{huang2019reachnn,fan2019towards,ivanov2019verisig,dutta2019reachability,tran2020cav_tool}, where a single planner is considered, our work addresses a hierarchical planner and thus considers a hybrid system. Specifically, we develop novel partition and union techniques across reachability analysis to improve efficiency and accuracy.
%at every computation step for enabling the verification of complex systems with multiple possible scenarios. 
%Our work is also related to the verification of neural-network controlled systems~\cite{huang2019reachnn,fan2019towards,ivanov2019verisig,dutta2019reachability,tran2020cav_tool} and also verify the systems by computing a bounded time reachable set. However, these techniques only focus on a single controller, which makes the system continuous, while we design and verify a hierarchical planner and the system is thus hybrid. Specifically, motivated by the idea of partitioning the initial set~\cite{wang2020energy}, we develop novel partition and union techniques across reachability analysis to improve efficiency and accuracy at every computation step.

\smallskip
\noindent
In summary, our work makes the following contributions:
\begin{myitemize}
\item With empirical study and theoretical analysis, we show that for those systems that may evolve into multiple physical scenarios, single neural network based planner is either unsafe, or extremely difficult to verify. 
\item We design a novel hierarchical neural network based planner with assured safety and better verifiability, based on the underlying physical scenarios of the system.
\item We develop novel partition and union techniques to improve efficiency and accuracy of reachability analysis, and propose an overapproximation method for the system under our hierarchical planner.
\item We demonstrate the safety enhancement from our hierarchical design through case studies of unprotected left turn and highway merging, compared with single neural network based planners. 
\end{myitemize}

The rest of the paper is organized as follows. Section~\ref{sec:formulation} introduces an illustrating example and defines the problem formulation. Section~\ref{sec:verification} presents our planner design and verification approach. Section~\ref{sec:experiment} shows the case studies and Section~\ref{sec:conclusion} concludes the paper.

\section{Problem Formulation}~\label{sec:formulation}

\begin{figure}[tbp]
\centering
\includegraphics[scale=0.32]{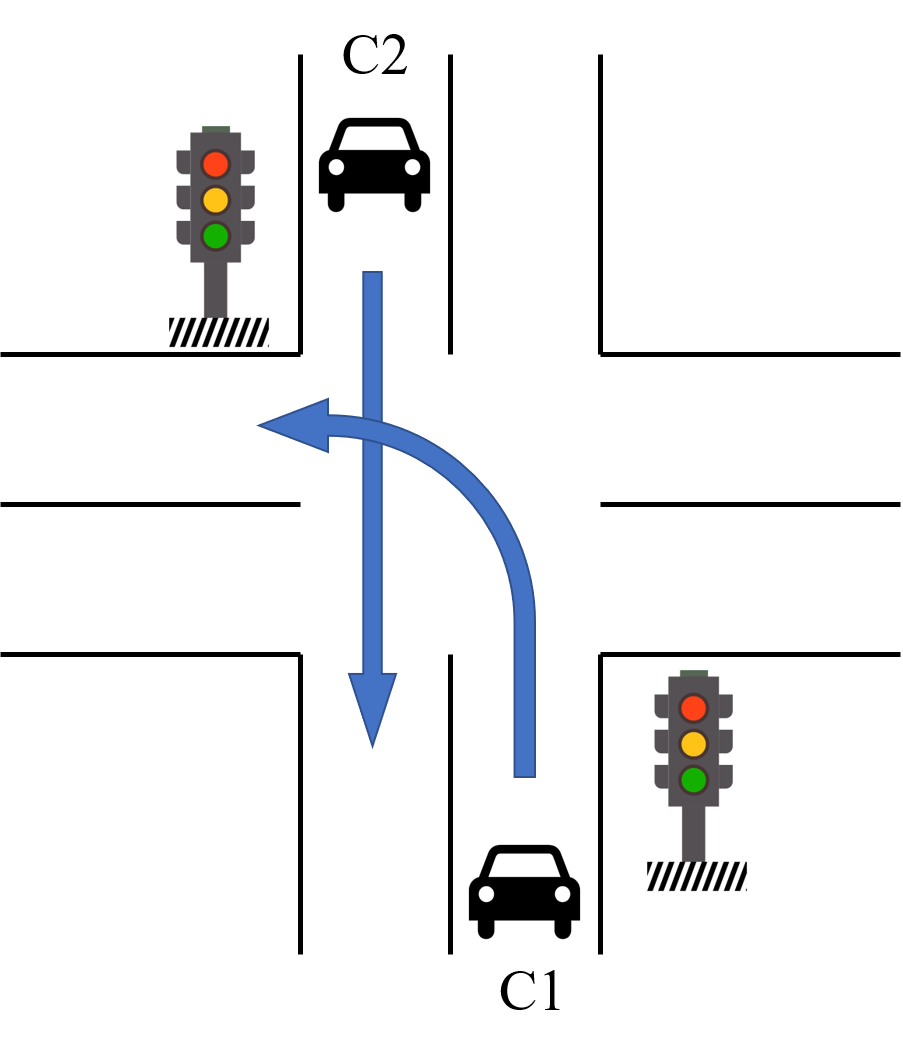}
\caption{The unprotected left turn system.
%Vehicle $C1$ is turning left and vehicle $C2$ is going straight from the opposite direction. Depending on the position and velocity of vehicle $C1$, estimated arriving time of vehicle $C2$ and the traffic signal, vehicle $C1$ may stop before the intersection, yield to vehicle $C2$, or proceed.
}
\label{fig:scenario}
%\vspace{-12pt}
\end{figure}

We consider the unprotected left turn as a representative application where different planning decisions and system states can eventually lead to different physical scenarios. The system includes a left turn vehicle $C1$ and another vehicle $C2$ going straight from the opposite direction in an intersection, as shown in Fig.~\ref{fig:scenario}. We model it as a 5-dimensional system:
\begin{equation} \label{eq:hybrid}
\begin{cases}
%\begin{aligned}
\dot p_1(t)=v_1(t)
\\
\dot v_1(t)=u(t)
\\
\dot \tau_{min}(t)=f_1(t)
\\
\dot \tau_{max}(t)=f_2(t)
\\
\dot t=1
%\end{aligned}
\end{cases}
\end{equation}
%\yixuan{define $f_1, f_2$ here} {\color{blue}I will do it later, we have $f_1=f_2=0$ first, then it changes later.} 
where $p_1(t)$ and $v_1(t)$ are the position and velocity of vehicle $C1$. Vehicle $C2$ is predicted to pass the conflicting area (where the two vehicles may potentially collide) in this intersection within the time window $\left[ \tau_{min}(t) \text{, } \tau_{max}(t) \right]$. $u(t)$ is the control input, representing the acceleration of vehicle $C1$. We assume that vehicle $C1$ follows a given path in the intersection to turn left, thus its trajectory can be derived with $u(t)$. $\tau_{min}(t)$ and $\tau_{max}(t)$ may change over time as vehicle $C1$ can update its prediction for $C2$. We assume that this time window will become tighter as two vehicles get closer to each other, i.e., $f_1(t)>=0$ and $f_2(t)<=0$. We also assume that the traffic signal follows a fixed pattern. First it is green for $t_g$ seconds, then it turns yellow for $t_y$ seconds, and then it turns red for $t_r$ seconds. After that, it turns back to green and repeats this turning pattern.
%The system can be modeled as a hybrid automata, in which transitions can be triggered by collision avoidance behavior of vehicle $C2$ or traffic signal changes. 

\begin{figure}[tbp]
\centering
\includegraphics[scale=0.48]{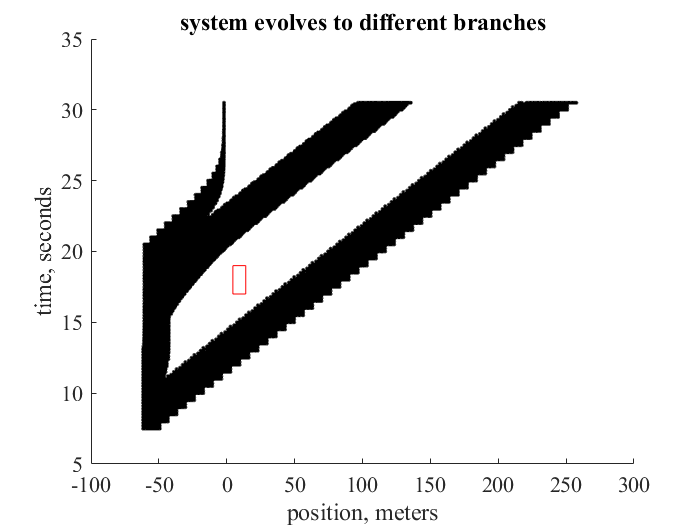}
\caption{System in Fig.~\ref{fig:scenario} evolves to multiple physical scenarios. The horizontal axis denotes the position of vehicle $C1$ along the planned path, and a negative value means that $C1$ has not entered the intersection. The vertical axis denotes time. The black region represents simulated trajectories of the system. From left to right, the three branches correspond to the scenarios where vehicle $C1$ stops before the intersection, yields to $C2$, and proceeds, respectively. The red rectangle is the unsafe region as vehicle $C2$ is expected to passing the intersection at time interval $\left[ 17 \text{, } 19 \right]$.}
\label{fig:branches}
\end{figure}

Fig.~\ref{fig:branches} shows the simulated trajectories based on human driving norm. The horizontal axis denotes the position of vehicle $C1$ along the planned path, where a negative value represents that $C1$ has not entered the intersection. When $p_1=4.5$ meters, $C1$ enters the region where two vehicles' paths may intersect. When $p_1=14$ meters, it leaves that conflicting region. There are three obvious branches as time goes on. The leftmost branch corresponds to the behavior of $C1$ stopping before the intersection, when it cannot pass the intersection before the signal turns red. The middle branch corresponds to the behavior of $C1$ yielding to vehicle $C2$ that goes straight, when there is potential danger for collision and $C2$ has the right of way. The rightmost branch corresponds to the behavior of $C1$ proceeding, when it is safe to pass the intersection before $C2$. The red rectangle marks the unsafe region, as vehicle $C2$ is expected to pass the conflicting region within the time window $\left[ \tau_{min}(t) \text{, } \tau_{max}(t) \right] \equiv \left[ 17 \text{, } 19 \right]$ seconds.

To react safely and efficiently, depending on the initial system state and changes in the surrounding traffic, vehicle $C1$ may take different actions. Note that although there may exist some planner $u'(t)$ that is safe and can lead to only one branch of system trajectories, i.e., braking and then stopping before the intersection in any case, it is not efficient and ideal in real life. 

%For this kind of complex systems, assuring safety and verification are very challenging. 

In some simpler systems such as adaptive cruise control and emergency braking~\cite{tran2019safety,huang2020opportunistic}, the system state may converge to a constant distance gap or gradually slows down to full stop. Here, the potential reachable states of the unprotected left turn system do not converge to a single scenario, but evolve to multiple different scenarios. This presents significant challenges to safety verification and assurance.

%states usually can converge well, i.e., maintaining a constant distance gap or gradually slow down to fully stop. Different from these systems, unprotected left turn does not have good convergence, it may even evolve to multiple different physical scenarios. 

%We consider different planner designs for vehicle $C1$, as detailed in section~\ref{sec:planner design}. 
Thus, in this work, we are interested in the following questions: Can vehicle $C1$ turn left safely and efficiently under our designed planner when facing different traffic scenarios, i.e., turning left without hesitance when it is safe and decelerating when facing potential collision? If so, can we formally verify the system safety under our designed planner? To answer these, we will first generally formulate the above-mentioned system where different planning decisions and system states can eventually lead to different physical scenarios.

\smallskip
\noindent\textbf{General Formulation.}
We consider a dynamic system:
\begin{equation} \label{eq:gen1}
\begin{cases}
%\begin{aligned}
\dot x(t)=f(x(t) \text{, } u(t))\text{, }\forall t \geq 0 \\
x(t) \in X-S_u \text{, } x(0) \in I \text{, } u(t) \in U 
%\end{aligned}
\end{cases}
\end{equation}
where $x(t)$ is the state variable, and $u(t)$ is the control input variable. We assume that $f$ is Lipschitz continuous in $x$ and continuous in $u$ to ensure the uniqueness of solution. By including time $t$ in state variable $x(t)$, system function $f$ can be time-variant. $X=\{ x\in {\mathbb{R}}^n \}$ is the state space. $S_u=\{ x\in {\mathbb{R}}^n | \underset{i}{\land} h_i(x) \leq 0 \}$ is the unsafe state space, $h$ denotes the linear constraint function and state $x$ is unsafe if $\underset{i}{\land} h_i(x) \leq 0$ is satisfied. $X-S_u$ is the set difference of $X$ and $S_u$. $I \subseteq X-S_u$ is the initial set of system state. %\chao{Need to emphasize the difference between the system here and the traditional control system, nonconvexity of the constraints, which will natually introduce the notion of branch}{\color{blue}it is stated in the paragraph above problem 1}. 
$U=\{ u\in {\mathbb{R}}^m \}$ is the control input space.

Let $\delta_c$ denote the control time stepsize. At time $t=i*\delta_c$, $i=0\text{, } 1 \text{, } 2 \text{, \ldots}$, the system controller $\kappa$ takes current state $x(i*\delta_c)$ and computes control input $u(i*\delta_c)=\kappa(x(i*\delta_c))$ for the next time step, the system becomes $\dot x(t)=f(x(t) \text{, } u(i*\delta_c))$ in the time interval $t \in \left[ i*\delta_c \text{, } (i+1)*\delta_c \right]$.

% \yixuan{How about introducing initial space and sampling period here?(They are part of the system.) Does the system have a goal to achieve, like left turn is successful for C1?}{\color{blue}initial space and sampling period are introduced. The system does have a goal, but it seems not easy to describe, and we want general description here, not restricted to unprotected left turn}

The trajectory $\varphi_{x(0)}$ to the system~\eqref{eq:gen1} starting from an initial state $x(0)$ can be formulated as:
\begin{equation} \label{eq:gen2}
%\begin{cases}
%\begin{aligned}
\dot \varphi_{x(0)}(t)=f(\varphi_{x(0)}(t) \text{, } u(t)) \text{, } x(0) \in I
%\end{aligned}
%\end{cases}
\end{equation}
where $\varphi_{x(0)}(0)=x(0)$.
%and $I$ is the initial state space, 
% \yixuan{is $I \subseteq X-S_u$?}.{\color{blue} yes, I add it in last equation's description}

With a well-designed controller $u(t)$, the system trajectories will evolve to disjoint subsets at step $i$ (and possibly at the following steps as well) to avoid the unsafe set. That is:
\begin{equation} \label{eq:gen3}
\begin{cases}
%\begin{aligned}
\varphi_{x(0)}(t) \in \underset{k}{\cup} S_k \text{, }\forall t \in \left[ (i-1)*\delta_c \text{, } i*\delta_c \right] \text{, }\forall x(0) \in I
\\
||x-x'|| > \epsilon_x \text{, } \exists \epsilon_x>0 \text{, }\forall x \in S_k \text{, } \forall x' \in S_j \text{, }\forall k \neq j 
%\\
%S_k \cap S_u = \emptyset \text{, }\forall k
%\end{aligned}
\end{cases}
\end{equation}
where $\delta_c$ is the control time stepsize, $S_k$ is a subset of system states in time interval $\left[ (i-1)*\delta_c \text{, } i*\delta_c \right]$, $\underset{k}{\cup} S_k$ is the union of all subsets. The distance between any two elements $x \in S_k$ and $x' \in S_j$ is always strictly greater than a positive real number $\epsilon_x$, for any two different subsets $S_k$ and $S_j$. 

It becomes more challenging to design a safety-assured planner due to the properties of the system as in~\eqref{eq:gen1} and \eqref{eq:gen3}. Since the safe state space $X-S_u$ is non-convex, the system reachable set needs to evolve to multiple branches to avoid the unsafe set. However, a planner $\kappa(x)$ that is Lipschitz continuous in $x$ intuitively cannot output significant different control signal $u(x)$ under only minor changes in system state $x$. For these complex systems, accidents cannot be prevented in experiments with previous neural network based planner designs, including hierarchical planners~\cite{naveed2020trajectory,cao2020reinforcement,nosrati2018towards,li2021safe,wang2020learning,ma2021model}. Thus, we try to answer: is there a planner design that can enable the change of system trajectory under changing scenarios? If so, can we verify its safety as the system reachable states evolve to multiple possible scenarios? Formally, we try to solve:

\begin{problem}
    For a dynamical system defined by~\eqref{eq:gen1} and~\eqref{eq:gen2}, is there a planner design $\kappa$ that can satisfy~\eqref{eq:gen3}?
\end{problem}

%Different from single neural network based planner, the hierarchical neural network based planner $\mu(\kappa_1 \text{, } \kappa_2 \text{, \ldots, } \kappa_N)$ is composed of one neural network selector $\mu$ and $N$ neural network planners $\kappa_i \text{, } i=1 \text{, } 2 \text{, \ldots, } N$. Here is a similar problem for this planner.

%In this paper, we also consider the naive single neural network based planner $\kappa_s$, which takes system states as inputs and directly outputs control variable $u(t)$. We are also interested in the question: will these two designs have any difference in safety or verifiability? Then we have the next problem.

\begin{problem}
    If there exists a planner $\kappa$ that can satisfy~\eqref{eq:gen3}, the safety verification problem is to determine whether the controlled trajectory $\varphi_{x(0)}(t) \in X-S_u \text{, }\forall t \geq 0 \text{, }\forall x(0) \in I$.
\end{problem}

\section{Planner Design and Safety Verification}\label{sec:verification}

In this section we first conduct formal analysis for  systems under a single neural network based planner. With theoretical analysis, we show that single neural network planners cannot handle well the systems that may evolve into different scenarios, and are hard to verify. To overcome these challenges, we present our design of a hierarchical neural network based planner, and then introduce the partition and union algorithm we developed for the verification of our hierarchical planner.

%cannot handle the systems that are expected to have branches of trajectories, and discuss the verifiability. To overcome these drawbacks of single planner, we present our hierarchical neural network based planner, finally present the partition \& union algorithm for hierarchical neural network planner verification.

%Then in this work we want to address the safety verification problems for the system with different planner designs, including single neural network and hierarchical neural network planner.
%$\kappa_s$, the other is for the system with scene decomposition based hierarchical neural network planner $\mu(\kappa_1 \text{, } \kappa_2 \text{, \ldots, } \kappa_N)$.

\subsection{Formal Analysis of Single Planner Design}

%\chao{Say single neural network planner is a popular design methodology by citing some papers}
Using a single neural network for planner design is well-studied~\cite{chen2020input,qureshi2020motion,seidman2020robust}. Deep neural networks provide better performance for complex systems than many traditional methods~\cite{dong2020facilitating,zhou2019development,zeng2019end}. However, single neural network based planner has its limitations, especially for safety-critical systems~\cite{cao2020reinforcement}. Below we formally analyze the system under a single neural network based planner with reachability analysis of a neural network controlled system~\cite{tran2020cav_tool,huang2019reachnn,ivanov2019verisig,dutta2019reachability}, and we leverage the Bernstein polynomial based reachability analysis~\cite{huang2019reachnn}, as it can handle neural networks with general and heterogeneous activation functions.
%based on reachability, which is a general property to analyze safety of a neural-network controlled system \cite{tran2020cav_tool,huang2019reachnn,ivanov2019verisig,dutta2019reachability}. In this paper, we consider to use Bernstein polynomial based approach in \cite{huang2019reachnn}, since it can handle neural networks with general and heterogeneous activation functions. 
Let us start with introducing reachable set and Bernstein polynomial.
\begin{definition}
A system state $x$ is reachable at time $t \geq 0$ on a system defined by~\eqref{eq:gen1} and \eqref{eq:gen2}, if and only if there exists $x(0) \in I$ such that $x=\varphi_{x(0)}(t)$. The reachable set $R$ of the system is defined as the set of all reachable states $R=\{x | x=\varphi_{x(0)}(t) \text{, } \forall t \geq 0 \text{, }\forall x(0) \in I \}$.
\end{definition}

The system is considered to be safe if its reachable set $R$ has no overlap with the unsafe set $S_u$. However, it is proven that computing the exact reachable set for most nonlinear systems is an undecidable problem~\cite{henzinger1998s}, not to mention systems with neural network planners. Thus, recent works mainly consider overapproximation of the reachable set. Safety can still be guaranteed if the overapproximated reachable set has no overlap with the unsafe set. 
%, while it is not assured if there is an overlap. It is 
Note that in this paper, for simplificy, we use the same notation for both the reachable set and its overapproximation. %, we use the same character.

For a controller/planner $\kappa_s$ defined over a n-dimensional state $x$, its Bernstein polynomials $B_{\kappa_s,d}(x)$ under degree $d=(d_1 \text{, } d_2 \text{, \ldots, } d_n)$ is:
\begin{equation} \label{eq:bern}
%\begin{cases}
\begin{aligned}
B_{\kappa_s,d}(x)= & \sum_{\substack{0 \leq a_j \leq d_j \\ j \in \{ 1 \text{, } 2 \text{, \ldots, } n\}}} \kappa_s \bigg(\frac{a_1}{d_1} \text{, } \frac{a_2}{d_2} \text{, \ldots, } \frac{a_n}{d_n} \bigg) 
\\&
\prod_{j=1}^{n} \bigg( \Big(\substack{d_j \\a_j}\Big)x^{a_j}_{j}(1-x_j)^{d_j-a_j} \bigg)
\end{aligned}
%\end{cases}
\end{equation}
where $\Big(\substack{d_j \\a_j}\Big)$ is a binomial coefficient.

To obtain an overapproximation of the reachable set for a system with a neural network based controller $\kappa_s$, we compute an overapproximation of the controller $\kappa_s$ using Bernstein polynomials similarly as in~\cite{huang2019reachnn}. Note that the reachable set is computed step by step and it is sufficient to perform the overapproximation of $\kappa_s$ at step $i$ on the latest computed reachable set $R_{i-1}$. 
%acquire overapproximation of controller $\kappa_s$ in a similar way as in~\cite{huang2019reachnn}. 
That is, $\kappa_s$ is overly approximated by a Bernstein polynomial with bounded error $\epsilon$ on set $R_{i-1}$ as: 
\begin{equation} \label{eq:err_bound}
%\begin{cases}
%\begin{aligned}
\kappa_s(x) \in B_{\kappa_s,d}(x) + \left[-\epsilon \text{, } \epsilon \right] \text{, } \forall x \in R_{i-1}
%\end{aligned}
%\end{cases}
\end{equation}
where $R_{0}=I$ when performing overapproximation in the first step. In the rest of the paper, $B_{\kappa_s}(x)$ is short for $B_{\kappa_s,d}(x)$, as it is not necessary to have the same $d$ for the overapproximation of different controllers.

With the above approach, the dynamic system with a single neural network based planner $\kappa_s$ is transformed into a polynomial system for computing the overapproximation of the reachable set. This enables our following analysis. 
%For more details, please refer to~\cite{huang2019reachnn}.

%\chao{Give some intuition why we need to introduce Lipschitz continuity and Proposition 3.1.}

\smallskip
\noindent\textbf{Challenge on correctness.} Intuitively it is unlikely that Lipschitz continuous planners can output significantly different control signal $u(x)$ under only minor changes in the system state $x$, and enable system trajectory go into several disjoint subsets under different scenarios. %How can they exactly control the system on a non-convex safe state space? 
Next, let us formally introduce Lipschitz constant and explain that a large number of single neural network planners may indeed be unsafe for the system.

\begin{definition}
A real-valued function $f:X \rightarrow \mathbb{R}$ is called Lipschitz continuous over $X\subseteq\mathbb{R}^n$, if there exists a non-negative real $L$, such that $||f(x)-f(x')||\leq L||x-x'||$ for $\forall x$, $x'\in X$. Any such $L$ is called a Lipschitz constant of $f$ over $X$.
\end{definition}

%Recent works present that a large number of neural networks are Lipschitz continuous and provide upper bound of Lipschitz constant~\cite{ruan2018reachability,huang2019reachnn}, such as convolutional or fully connected neural networks with ReLU, sigmoid and hyperbolic tangent (tanh) activation functions.

%{\color{blue}However, as revealed later by experimental results in section~\ref{sec:experiment}, in the case that a system will evolve into several branches to avoid the unsafe set, single neural network based planner $\kappa_s$ probably is unsafe, if not, it is hard to verify the safety. }

\begin{figure*}[tbp]
\centering
\includegraphics[scale=0.43]{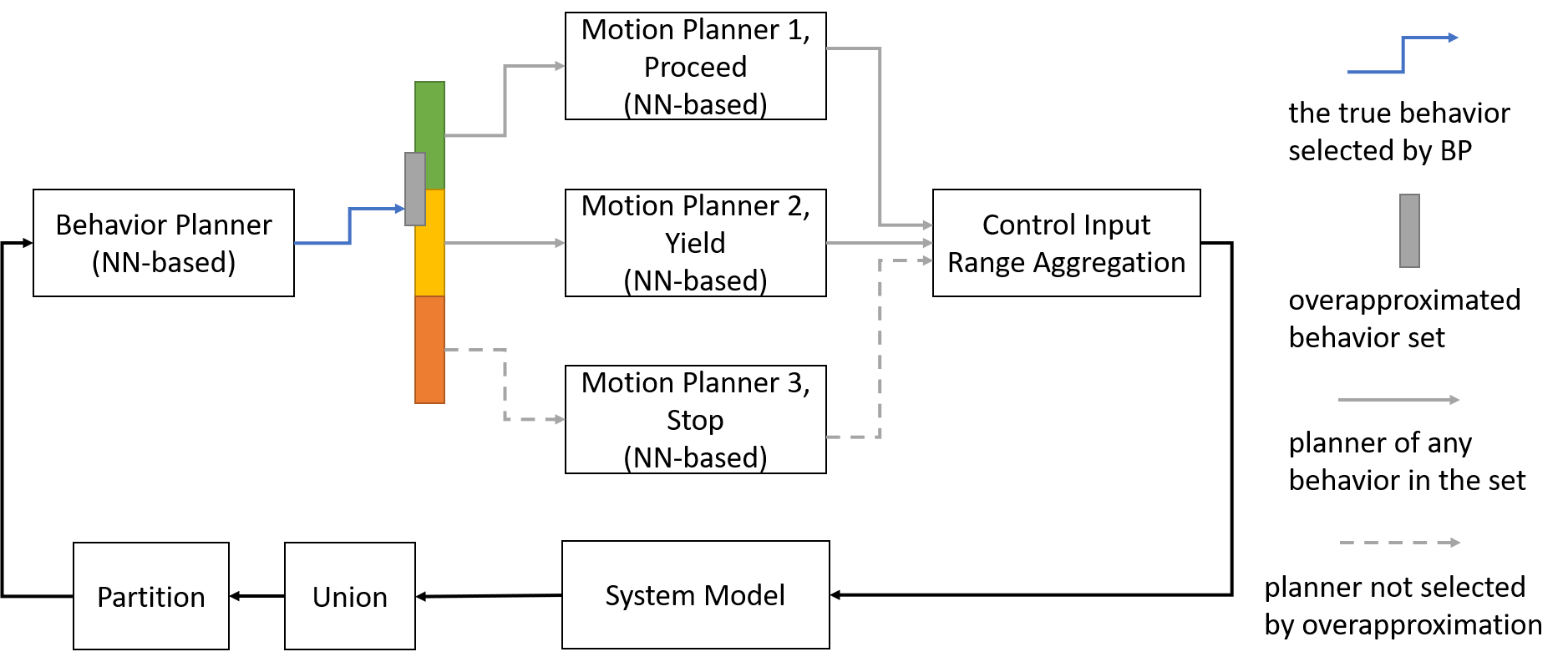}
\caption{Our design of the hierarchical neural network based planner consists of one behavior planner $\mu$ and $N$ motion planners $\{\kappa_1 \text{, } \kappa_2 \text{, \ldots, } \kappa_N\}$. Take the unprotected left turn system as an example, there are three underlying physical scenarios: vehicle $C1$ may stop before the intersection, yield to vehicle $C2$, or proceed, and they correspond to three motion planners shown here in the figure. The behavior planner decides the most appropriate behavior for vehicle $C1$ given the system state $x$, and then the corresponding motion planner is enabled to control the system. To compute an overapproximation of the reachable set of the system under such hierarchical planner, we first compute an overapproximated behavior set with Bernstein polynomial approximation as in Eq.~\eqref{eq:bern} and \eqref{eq:err_bound}, which is illustrated by the grey rectangle in the figure. Then for each behavior in the overapproximated behavior set, the corresponding motion planner's output range can be aggregated as the possible control input range, thus computing an overapproximation of the system state reachable set under all possible behaviors.
%The Partition and Union module can improve the efficiency and accuracy of such reachable set computation, as detailed in Algorithm~\ref{contract:partition and union}. For other applications, the number of motion planners could be different, e.g., there are only proceed and yield planners in the case of highway merging.
}
\label{fig:layered}
\end{figure*}

\begin{proposition}\label{proposition1}
For a dynamical system defined by~\eqref{eq:gen1} and \eqref{eq:gen2} with single neural network based planner $\kappa_s$, if $\kappa_s$ is a convolutional or fully connected neural network with ReLU, sigmoid or hyperbolic tangent (tanh) activation functions, then the controlled trajectory $\varphi_{x'(0)}(t')$ will not evolve to several branches as formulated in \eqref{eq:gen3}.
\end{proposition}

\begin{proof}
We will first prove that if a neural network planner $\kappa_s$ can ensure that the system evolves to several branches as defined in~\eqref{eq:gen3}, $\kappa_s$ is not Lipschitz continuous. We assume it is at step $i$ that the reachable set $R_i$ can be represented as $R_i=\underset{k}{\cup} S_{k}$ as in~\eqref{eq:gen3} for the first time. Then there exists $x_{i-1} \in R_{i-1}$ and $x'_{i-1}\in R_{i-1}$ such that $||x_{i-1}-x'_{i-1}||\rightarrow 0$, $x_{i-1}+f(x_{i-1}\text{, }\kappa_s(x_{i-1}))*\delta t=x_i \in R_k$, $x'_{i-1}+f(x'_{i-1}\text{, }\kappa_s(x'_{i-1}))*\delta t=x'_i \in R_j$, $k \neq j$ and $\delta t \rightarrow 0$. According to~\eqref{eq:gen3}, there exists $\epsilon_x>0$ such that $||x_i-x'_i||>\epsilon_x$, and we have $||f(x_{i-1}\text{, }\kappa_s(x_{i-1}))-f(x'_{i-1}\text{, }\kappa_s(x'_{i-1}))||>\epsilon_f$ and $\epsilon_f>0$. Because $f$ is Lipschitz continuous, then $\kappa_s$ is not Lipschitz continuous. However, since $\kappa_s$ is a convolutional or fully connected neural network with ReLU, sigmoid and hyperbolic tangent (tanh) activation functions, it should be Lipschitz continuous~\cite{huang2019reachnn}. From this contradiction, we know that the controlled trajectory $\varphi_{x'(0)}(t')$ will not evolve to several branches as formulated in \eqref{eq:gen3}.
\end{proof}

Based on this proposition, for most single neural network based planners, the system reachable set will cover the unsafe region $S_u$ and there is actually no disjoint subsets. 

\smallskip
\noindent\textbf{Challenge on verifiability.} %\chao{Could we clarify this point more detail?} 
Even if $\kappa_s$ is a neural network based planner that can satisfy~\eqref{eq:gen3}, it will have infinitely large Lipschitz constant according to Proposition~\ref{proposition1}. This typically makes the safety verification extremely hard due to the importance of Lipschitz constant in the construction of reachable sets. As observed in~\cite{fan2019towards, wang2021cocktail, ivanov2019verisig} and shown in our case studies in Sections~\ref{sec:experiment single} and~\ref{sec:experiment single_merging}, the reachable set expands more quickly when the Lipschitz constant of neural network based planner is larger. In which case, the verification process may terminate due to uncontrollable approximation error or excessively long computation time.
These challenges can be overcome in our hierarchical planner design as introduced below.

%In contrast, these difficulties can be easily prevented under our hierarchical planner composed of all Lipschitz continuous neural networks, which will be introduced below.

\subsection{Hierarchical Planner Design and Reachability Analysis}\label{sec:planner design}

\noindent\textbf{Hierarchical planner design.} The drawbacks of a single neural network based planner motivates us to propose a hierarchical planner, as shown in Fig.~\ref{fig:layered}. The main idea is to learn different motion planners for different physical scenarios and a system-level behavior planner for changing between scenarios. Specifically, our hierarchical planner consists of a behavior planner $\mu$ and $N$ motion planners $\{\kappa_1 \text{, } \kappa_2 \text{, \ldots, } \kappa_N\}$, assuming that the system may evolve into $N$ different physical scenarios. These planners are all neural network based and take system states as inputs. The behavior planner's output $\mu(x)$ can be mapped to the discrete behavior choice by mapping function $f_m$ and we have $f_m(\mu(x)) \in \{\kappa_1 \text{, } \kappa_2 \text{, \ldots, } \kappa_N\}$, while motion planners output control variable $u(t)$. To illustrate the idea, we still use the unprotected left turn system as an example: The behavior planner will decide the most appropriate behavior for vehicle $C1$ given the system state $x$. Then the corresponding motion planner, e.g., motion planner $2$ in Fig.~\ref{fig:layered}, will be enabled to control the system. The system will use the same motion planner before the next triggering of the behavior planner. Note that it is flexible to set the trigger conditions for the behavior planner, e.g., behavior planner may be triggered every $t_{bp}$ seconds when the system state has significant changes.  %We denote this hierarchical planner by $\mu(\kappa_1 \text{, } \kappa_2 \text{, \ldots, } \kappa_N)$. %Then the first problem we want to address is the safety verification of the system with planner $\mu(\kappa_1 \text{, } \kappa_2 \text{, \ldots, } \kappa_N)$, formulated as below.

%\chao{Mainly discuss here why the hierarchical planner could be more "correct"} 
The challenges under a single neural network planner $\kappa_s$ can be overcome in our hierarchical planner design as the reachable sets computed under different motion planners correspond to different physical scenarios and are disjoint. For each underlying physical scenario, the corresponding motion planner can generate system trajectory to avoid the unsafe region. Given that all motion planners are safe in their corresponding scenarios, system safety can be guaranteed with the computation of an overapproximated behavior set for the system-level behavior planner.  
\smallskip
\noindent\textbf{Reachability verification.} Next we first present our partition and union algorithm for general reachable set computation to improve efficiency and accuracy, and then we introduce our method to overapproximate the reachable set for a system under a hierarchical planner $\mu(\kappa_1 \text{, } \kappa_2 \text{, \ldots, } \kappa_N)$.

\begin{algorithm}[t]
\SetAlgoLined
\KwResult{Reachable set $R$ and verification result %\chao{Looks good. Indicate the line number in the paragraph. Like "We first do .... (line *)"}{\color{blue}how is it now?}
}
\SetKwInOut{Input}{Input}
\Input{Initial set $I$, unsafe set $S_u$ and goal set $S_g$}
 $R \leftarrow I$\;
 \While{$I-S_g \neq \emptyset$ and $R \cap S_u = \emptyset$}{
 %Partition set $I$ into grids of size $\delta$\;
 $\{I_\gamma\} \leftarrow Partition(I \text{, } \delta)$\;
 \For{each grid $I_\gamma$}{
 %Compute reachable set $R_\gamma$ in next $n$ steps\;
 $R_\gamma \leftarrow Reach\_Comp(I_\gamma \text{, } n)$\;
 %approximation error $\epsilon > e$ is larger than threshold $e$, memory, exceed maximum time
 \While{$Reach\_Comp$ terminates without returning $R_\gamma$}{
 %Further partition set $I_\gamma$ and compute reachable set\;
 $\{I_{\gamma'}\} \leftarrow Partition(I_\gamma \text{, } \delta /2)$\;
 \For{each grid $I_{\gamma'}$}{
 $R_{\gamma'} \leftarrow Reach\_Comp(I_{\gamma'} \text{, } n)$\;
 }
 }
 }
 %Reset set $I$ as the union of all computed reachable sets $R_\gamma$\;
 $I \leftarrow \underset{\gamma}{\cup} R_{\gamma}$\;
 %Append set $I$ to set $R$\;
 $R \leftarrow  R \cup I$\;
 }
 \eIf{$R \cap S_u = \emptyset$}{
 Verification result is safe.\
 }{
 Verification result is uncertain.\
 }
 \caption{Partition and Union for Reachable Set Computation}
 \label{contract:partition and union}
\end{algorithm}

%Motivated by the idea of partitioning the initial set, 
We develop a partition and union method that can improve the efficiency and accuracy of the overapproximated reachable set at every computation step, as shown in Algorithm~\ref{contract:partition and union}. Specifically, when the system state has not fully reached the goal set $S_g$ and the system is currently safe (line 2), we will keep computing the reachable set as follows. We first partition the initial set $I$ into grids of size $\delta$ (line 3), and then compute the reachable set for each grid $I_\gamma$ in the next $n$ steps (line 5). The computation process may terminate without returning $R_\gamma$ (line 6) due to memory limitation, low accuracy or no result within certain time. In that case, we will further partition the initial set $I_\gamma$ (line 7) and compute the reachable set (line 9). Once the reachable set $R_\gamma$ is computed for each grid, we union them as the system reachable set for this round, and reset the initial set $I=\underset{\gamma}{\cup} R_{\gamma}$ (line 13). This process repeats until the system state reaches the goal set $S_g$ and is verified to be safe, or the overapproximation of the reachable set has overlap with the unsafe set $S_u$ (which presents an uncertain verification results given the nature of overapproximation).

To compute an overapproximated reachable set for a system with planner $\mu(\kappa_1 \text{, } \kappa_2 \text{, \ldots, } \kappa_N)$, we first overapproximate $\mu(x)$ in a similar way as in~\eqref{eq:err_bound} as:
\begin{equation} \label{eq:err_mu}
%\begin{cases}
%\begin{aligned}
\mu(x) \in B_{\mu}(x) + \left[-\epsilon_{\mu} \text{, } \epsilon_{\mu} \right] \text{, } \forall x \in R_{i-1}
%\end{aligned}
%\end{cases}
\end{equation}
where $R_{i-1}$ is the overapproximated reachable set of the system in the $(i-1)$-th step. We then compute the overapproximated output range of $\mu(x)$, $R_{i}^{\mu}$, on the set $R_{i-1}$. By the mapping function $f_m$, we have the overapproximated set of the selected planner $S_{ctrl} \subseteq \{\kappa_1 \text{, } \kappa_2 \text{, \ldots, } \kappa_N\}$. For any planner $\kappa_a(x) \in S_{ctrl}$, we can compute its overapproximated system state reachable set $R^{\kappa_a}_i$ on $R_{i-1}$. Then, the overapproximated reachable set of the system at step $i$ is $R_i=\underset{\kappa_a \in S_{ctrl}}{\cup} R^{\kappa_a}_i$. The soundness of our approach can be proven below.

\begin{proposition}\label{proposition2}
%\begin{lemma}
(Soundness). For a dynamical system defined by~\eqref{eq:gen1}, \eqref{eq:gen2} and \eqref{eq:gen3} with a hierarchical neural network based planner $\mu(\kappa_1 \text{, } \kappa_2 \text{, \ldots, } \kappa_N)$, the controlled trajectory $\varphi_{x(0)}(t)$ satisfies $\varphi_{x(0)}(t) \in \underset{\kappa_a \in S_{ctrl}}{\cup} R^{\kappa_a}_i \text{, }\forall t \in \left[ (i-1)*\delta_c \text{, } i*\delta_c \right] \text{, }\forall i \in \mathbb{N}^{+} \text{, }\forall x(0) \in I$.
%\end{lemma}
\end{proposition}

\begin{proof}
Let us prove by contradiction. We assume that it is at step $i$ that we compute the wrong reachable set $R_i=\underset{\kappa_a \in S_{ctrl}}{\cup} R^{\kappa_a}_i$ for the first time. Thus $\exists x'(0) \in I$ and $\exists \kappa' \in \{ \kappa_1 \text{, } \kappa_2 \text{, \ldots, } \kappa_N\}$ such that $\varphi_{x'(0)}((i-1)*\delta_c)=x' \in R_{i-1}$, $\varphi_{x'(0)}(t) \notin \underset{\kappa_a \in S_{ctrl}}{\cup} R^{\kappa_a}_i$ and $f_m(\mu(x'))=\kappa'$. Since $f_m(\mu(x'))=\kappa'$, we have $\kappa' \in S_{ctrl}$ and $R^{\kappa'}_i \subseteq \underset{\kappa_a \in S_{ctrl}}{\cup} R^{\kappa_a}_i$. Finally $\varphi_{x'(0)}(t) \in R^{\kappa'}_i$ contradicts that $\varphi_{x'(0)}(t) \notin \underset{\kappa_a \in S_{ctrl}}{\cup} R^{\kappa_a}_i$.

% \yixuan{is it $\varphi_{x'(0)}(t)$ or $\varphi_{x'(0)}(i\times \delta)$?}
\end{proof}

\begin{figure}[t!]
\centering
\includegraphics[scale=0.48]{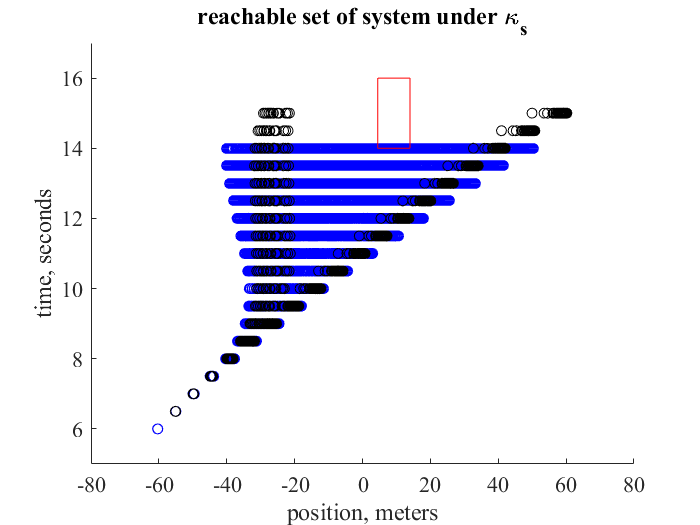}
\caption{Reachable set and sampled trajectories for the unprotected left turn system with a single neural network planner $\kappa_s$. The blue region is the overapproximated reachable set and the black region is 100 sampled trajectories from the same initial set. Initial set $I$ is set as $I=\{x \in \mathbb{R}^5|p_1\in \left[-60.4 \text{, } -60.3\right]\text{, } v_1\in \left[10.5 \text{, } 10.51\right]\text{, } \tau_{min}\equiv14\text{, } \tau_{max}\equiv16\text{, } t=6 \}$. Unsafe set is $S_u=\{x \in \mathbb{R}^5|p_1\in \left[4.5 \text{, } 14\right]\text{, } t \in \left[14 \text{, } 16\right] \}$, and it is marked with a red rectangle. }
\label{fig:reachx1}
\end{figure}

\begin{figure*}[t!]
\centering
\includegraphics[scale=0.42]{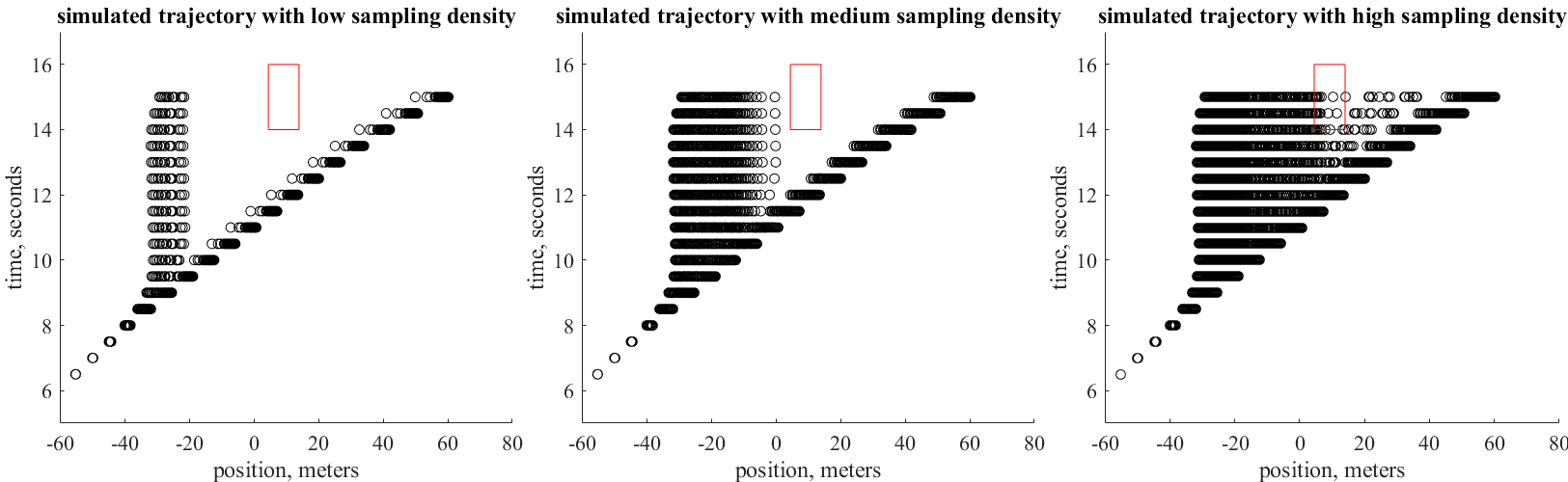}
\caption{Simulated trajectories with different sampling densities from the initial set $I$, for the unprotected left turn system with a single neural network based planner $\kappa_s$. The black region represents simulated trajectories and the red rectangle is the unsafe set $S_u$. $I$ and $S_u$ are the same as in Fig.~\ref{fig:reachx1}. From left to right, these three subplots correspond to the trajectories of one hundred, ten thousand and one million samples from the initial set $I$, respectively.}
\label{fig:sample}
\end{figure*}

\section{Case Studies}\label{sec:experiment}

Our hierarchical neural network based planner design and safety verification method can be applied to many safety critical systems defined by~\eqref{eq:gen1},~\eqref{eq:gen2} and~\eqref{eq:gen3}. In this section, we demonstrate its effectiveness in two case studies of unprotected left turn and highway merging in autonomous driving.

For both applications, we generate the trajectory dataset based on human driving norm and use the same dataset to train all the planning neural networks ($\kappa_s \text{, } \mu \text{, } \kappa_1 \text{, } \kappa_2 \text{, \ldots, } \kappa_N$). These neural networks all have two hidden layers, with each layer having ten neurons. We select ReLU and tanh as the activation functions for the hidden layer and the output layer, respectively. We set control time stepsize $\delta_c=0.5$ seconds for all experiments in this work. Based on the verification tool ReachNN~\cite{huang2019reachnn} and POLAR~\cite{huang2021polar}, we implement Algorithm~\ref{contract:partition and union} to compute the overapproximation of reachable set for the system under the single neural network based planner and the hierarchical neural network based planner, respectively. 

\subsection{Unprotected Left Turn}
We conduct experiments for the unprotected left turn system as described earlier in Section~\ref{sec:formulation}.

\subsubsection{Empirical Study of Single Neural Network Planner}\label{sec:experiment single}

Fig.~\ref{fig:reachx1} shows the overapproximated reachable set under a single neural network based planner $\kappa_s$, which is consistent with our analysis in Proposition~\ref{proposition1}. We select the initial set $I=\{x \in \mathbb{R}^5|p_1\in \left[-60.4 \text{, } -60.3\right]\text{, } v_1\in \left[10.5 \text{, } 10.51\right]\text{, } \tau_{min}\equiv 14\text{, } \tau_{max}\equiv 16\text{, } t=6 \}$ to cover different behaviors (proceed and yield) of vehicle $C1$. 
%We assume the motion estimations for the other vehicle $C2$ do not change, i.e., $\dot \tau_{min}=0$ and $\dot \tau_{max}=0$. 
The unsafe set $S_u$ is determined by $\tau_{min}$ and $\tau_{max}$, i.e., $S_u=\{x \in \mathbb{R}^5|p_1\in \left[4.5 \text{, } 14\right]\text{, } t \in \left[14 \text{, } 16\right] \}$. In this figure, the blue region represents the reachable set and the black region is the simulated trajectories of 100 sampling states from the same initial set. The reachable set cannot be represented with several disjoint subsets as described in Eq.~\eqref{eq:gen3} and it overlaps with the unsafe set $S_u$.

By increasing the number of sampling states from the same initial set $I$, we actually find counterexamples that prove $\kappa_s$ is indeed unsafe. Fig.~\ref{fig:sample} shows the simulated trajectories with different sampling density from $I$. The three subplots from left to right correspond to the trajectories of one hundred, ten thousand and one million samples in the initial set $I$, respectively. Based on our analysis in Proposition~\ref{proposition1}, for any other neural network planner $\kappa'_s$ that is a convolutional or fully connected neural network with ReLU, sigmoid and tanh activation functions, we can always find counterexamples by increasing the sampling density.

\subsubsection{Experiment Results of Hierarchical Planner}
\begin{figure*}[t!]
\centering
\includegraphics[scale=0.42]{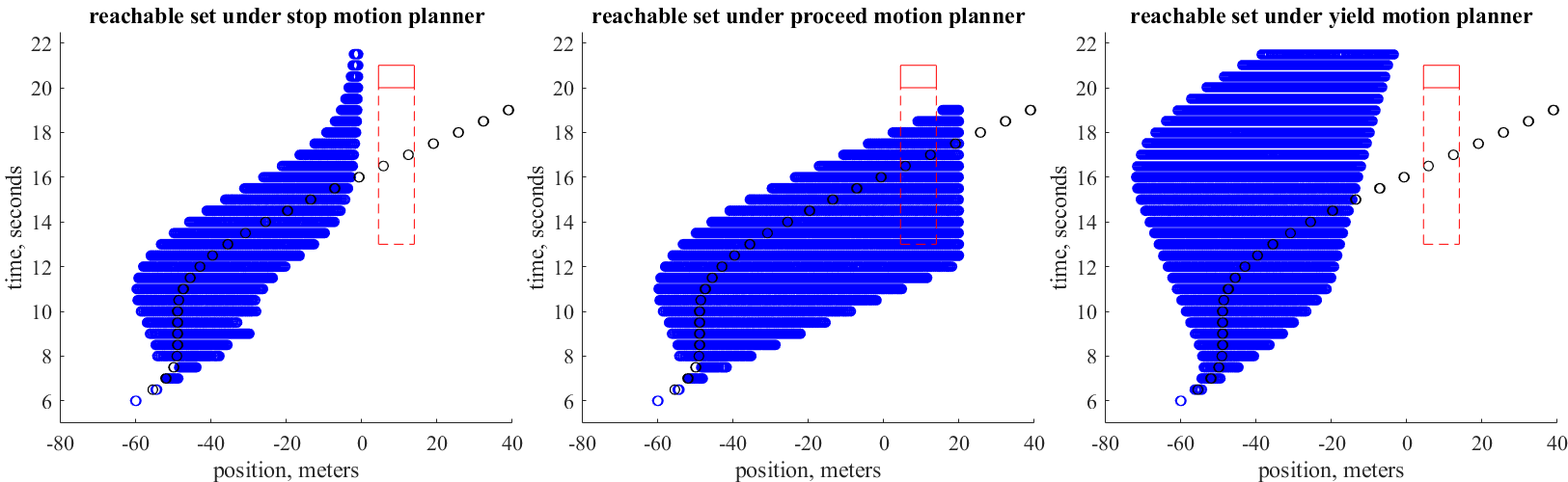}
\caption{Reachable set and sampled trajectories for the unprotected left turn system under hierarchical neural network planner with changing time window $\left[\tau_{min} \text{, } \tau_{max} \right]$. From left to right, the three subplots present the reachable system states under the stop motion planner, proceed motion planner, and yield motion planner respectively. The blue region is the overapproximated reachable set and the black region is 100 sampled trajectories from the same initial set. Initial set $I$ is set as $I=\{x \in \mathbb{R}^5|p_1\in \left[-60 \text{, } -59.7\right]\text{, } v_1\in \left[10.5 \text{, } 10.51\right]\text{, } \tau_{min}=13\text{, } \tau_{max}=21\text{, } t=6 \}$. The time window $\left[\tau_{min} \text{, } \tau_{max} \right]$ is initially $\left[13 \text{, } 21 \right]$ at time $6 \leq t < 7$, then $\left[15 \text{, } 21 \right]$ at time $7 \leq t < 8$, $\left[17 \text{, } 21 \right]$ at time $8 \leq t < 9$, $\left[19 \text{, } 21 \right]$ at time $9 \leq t < 10$, and finally $\left[20 \text{, } 21 \right]$ at time $t \geq 10$. The unsafe set $S_u$ changes with the time window $\left[\tau_{min} \text{, } \tau_{max} \right]$ as time goes on. It is initially the red rectangle with the dashed line and finally the red rectangle with the solid line. %The behavior planner is triggered when the time window changes, and thus the reachable set includes the trajectories that switches motion planners in the meantime. %It is noted that we consider $C1$ has already passed the intersection and is safe if the position is larger than 20 meters, thus system states with position of larger than 20 are excluded for reachable set computation in next step in the middle subplot.
}
\label{fig:reachx3}
\end{figure*}

\begin{figure}[htbp]
\centering
\includegraphics[scale=0.48]{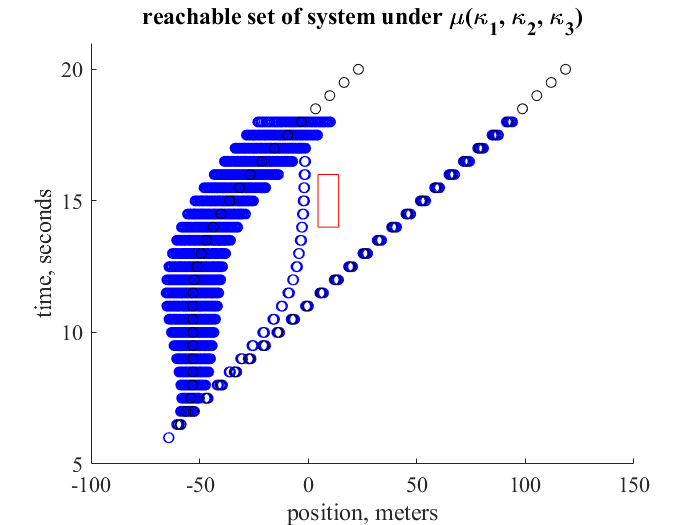}
\caption{Reachable set and sampled trajectories for the unprotected left turn system with hierarchical neural network planner $\mu(\kappa_1 \text{, } \kappa_2 \text{, } \kappa_3)$. The blue region is the overapproximated reachable set and the black region is 100 sampled trajectories from the same initial set. Initial set $I$ is set as $I=\{x \in \mathbb{R}^5|p_1\in \left[-64.35 \text{, } -64.05\right]\text{, } v_1\in \left[10.5 \text{, } 10.51\right]\text{, } \tau_{min}\equiv 14\text{, } \tau_{max}\equiv 16\text{, } t=6 \}$. Unsafe set is $S_u=\{x \in \mathbb{R}^5|p_1\in \left[4.5 \text{, } 14\right]\text{, } t \in \left[14 \text{, } 16\right] \}$, and it is marked with a red rectangle. States in the initial set $I$ is assessed by the behavior planner $\mu$ and in this example $S_{ctrl}$ includes all three behaviors (proceed, yield and stop). At the following time steps, vehicle $C1$ will not adjust its behavior and the reachable set of different behaviors are computed independently.}
\label{fig:reachx2}
\end{figure}

For the system with our design of a hierarchical planner $\mu(\kappa_1 \text{, } \kappa_2 \text{, } \kappa_3)$, we compute the overapproximation of the reachable set with the method introduced in Section~\ref{sec:planner design}, and the results are shown in Figs.~\ref{fig:reachx2} and ~\ref{fig:reachx3}.

First, we assume that $\tau_{min}$ and $\tau_{max}$ do not change over time. We set the initial set $I=\{x \in \mathbb{R}^5|p_1\in \left[-64.35 \text{, } -64.05\right]\text{, } v_1\in \left[10.5 \text{, } 10.51\right]\text{, } \tau_{min}\equiv 14\text{, } \tau_{max}\equiv 16\text{, } t=6 \}$ to cover different behaviors (proceed and yield) of vehicle $C1$. Then the unsafe set is $S_u=\{x \in \mathbb{R}^5|p_1\in \left[4.5 \text{, } 14\right]\text{, } t \in \left[14 \text{, } 16\right] \}$. As shown in Fig.~\ref{fig:reachx2}, due to overapproximation, all three behaviors are included in the reachable set $S_{ctrl}$ by the behavior planner $\mu$ at the beginning. Since the motion estimation for the other vehicle $C2$ remains the same, we assume that the behavior planner is triggered only once. % and vehicle $C1$ will not adjust its behavior in the mean time. 
From the figure, we can observe three disjoint reachable subsets in the time interval $t \in \left[10 \text{, } 16\right]$ and the reachable set has no overlap with the red unsafe region. Thus the planner $\mu(\kappa_1 \text{, } \kappa_2 \text{, } \kappa_3)$ is verified to be safe in this example.

We then consider the case where vehicle $C1$ updates the motion estimation for vehicle $C2$ as the two vehicles get closer to the intersection (which is often the case in practice), and we verify the system safety with the same hierarchical planner. Fig.~\ref{fig:reachx3} presents the system state reachable sets under different motion planners $\{R^{\kappa_1} \text{, } R^{\kappa_2} \text{, } R^{\kappa_3}\}$, which all together form the reachable set $R$ under our hierarchical planner. We set the initial set $I=\{x \in \mathbb{R}^5|p_1\in \left[-60 \text{, } -59.7\right]\text{, } v_1\in \left[10.5 \text{, } 10.51\right]\text{, } \tau_{min}=13\text{, } \tau_{max}=21\text{, } t=6 \}$. The time window $\left[\tau_{min} \text{, } \tau_{max} \right]$ is initially $\left[13 \text{, } 21 \right]$ at time $6 \leq t < 7$, then $\left[15 \text{, } 21 \right]$ at time $7 \leq t < 8$, $\left[17 \text{, } 21 \right]$ at time $8 \leq t < 9$, $\left[19 \text{, } 21 \right]$ at time $9 \leq t < 10$, and finally $\left[20 \text{, } 21 \right]$ at time $t \geq 10$. The unsafe set $S_u$ changes with the time window $\left[\tau_{min} \text{, } \tau_{max} \right]$ as time goes on. In Fig.~\ref{fig:reachx3}, the red rectangle with dashed line is the initial unsafe region, and the red rectangle with solid line is the final unsafe region. As there is no overlap between the reachable set with unsafe region, the planner $\mu(\kappa_1 \text{, } \kappa_2 \text{, } \kappa_3)$ is verified to be safe in this case where $C1$ updates its estimation on $C2$ and the time window $\left[\tau_{min} \text{, } \tau_{max} \right]$ is updated over time.

%There is an efficiency and verifiability trade-off relationship with the trigger condition of behavior planner. 

Since the behavior planner is triggered multiple times in the case in Fig.~\ref{fig:reachx3}, the eventual reachable set $R$ includes system trajectories under switched motion planners and is significantly larger than the reachable set in Fig.~\ref{fig:reachx2}. 
%in figure~\ref{fig:reachx3}, thus the set is significantly larger than the reachable set in figure~\ref{fig:reachx2}. 
Intuitively, if the behavior planner is triggered more frequently, the vehicle $C1$ can adapt to a more appropriate behavior sooner, but this will also results in a larger reachable set and more verification effort. 
%it is expected that vehicle $C1$ can adapt to a more appropriate behavior sooner. However, it also results in a larger reachable set and more verification effort.

\subsection{Highway Merging}

\begin{figure}[htbp]
\centering
\includegraphics[scale=0.45]{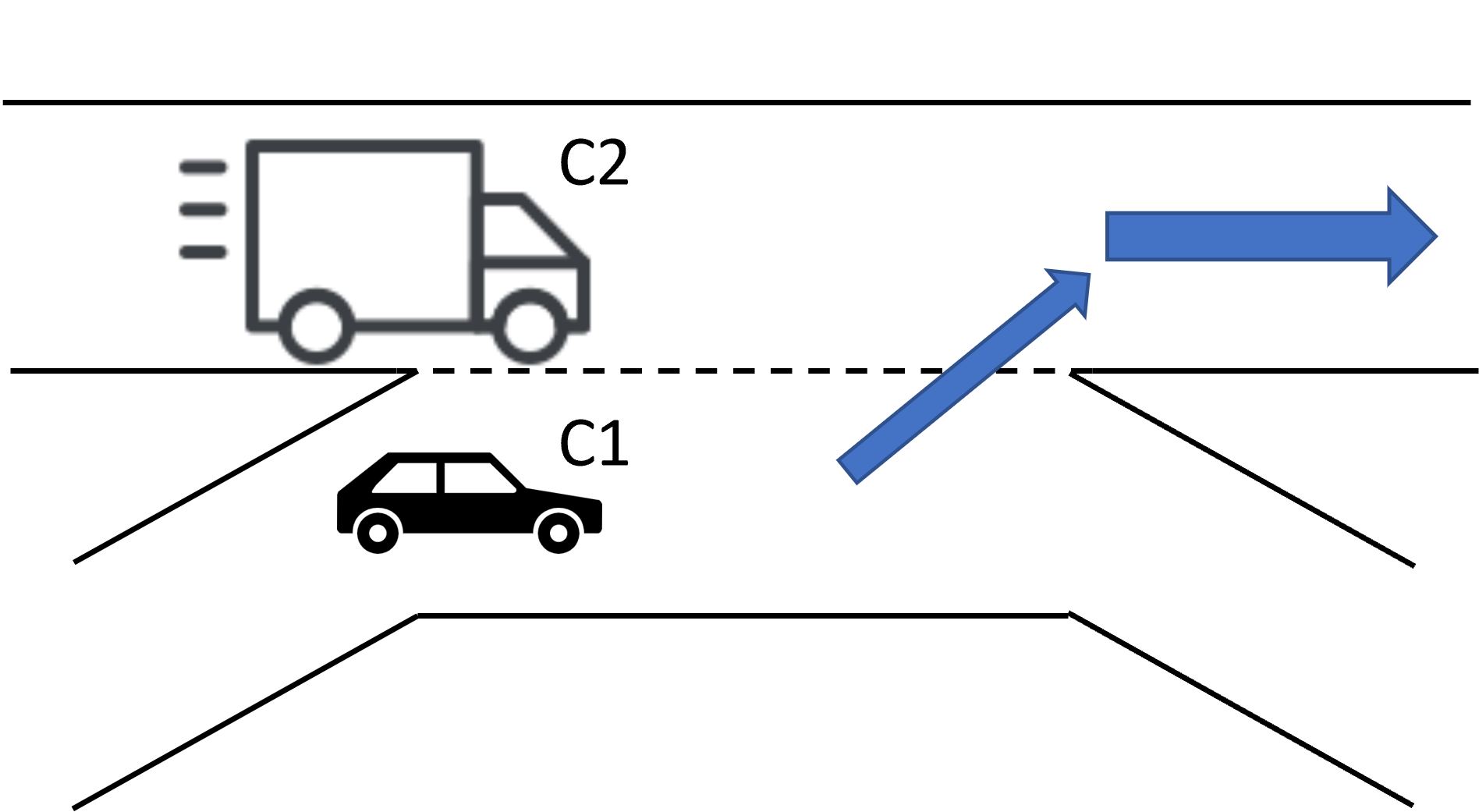}
\caption{The highway merging system. Vehicle $C1$ is merging onto the highway and vehicle $C2$ stays on the highway. Depending on the positions and velocities of vehicle $C1$ and $C2$, vehicle $C1$ may yield to vehicle $C2$, or proceed.}
\label{fig:merging}
\end{figure}

Another common and challenging task for autonomous driving is highway merging. As shown in Fig.~\ref{fig:merging}, vehicle $C1$ intends to merge onto the highway from on-ramp while another vehicle $C2$ stays on the highway. The system can be modeled as follows:
\begin{equation} \label{eq:hybrid_merging}
\begin{cases}
%\begin{aligned}
\dot p_1(t)=v_1(t)
\\
\dot v_1(t)=u(t)
\\
\dot p_2(t)=v_2(t)
\\
\dot v_2(t)=0
%\end{aligned}
\end{cases}
\end{equation}
where $p_1(t)$, $v_1(t)$, $p_2(t)$ and $v_2(t)$ are the longitudinal position and the velocity of vehicle $C1$ and $C2$, respectively. $u(t)$ is the control input, which is the acceleration of vehicle $C1$. In this example, we assume that vehicle $C2$ is a heavy truck and will maintain its speed.

To simplify this problem, we only discuss longitudinal motion of vehicle $C1$. Merging is considered to be feasible and safe if the longitudinal distance $| p_1(t_x) - p_2(t_x)|$ is larger than a threshold $d_{th}$ at some time point $t_x$ when vehicle $C1$ has not reached the end of the side road, i.e, $p_1(t_x)<p_{end}=150$ meters. Different from the unprotected left turn case, the unsafe set $S_u$ is not fixed here. The system is safe as long as there exists a position window $\left[p_{1,min} \text{, } p_{1,max}\right]$ for vehicle $C1$ and unsafe set $S_u=\{x \in \mathbb{R}^4|| p_1(t) - p_2(t)| \leq d_{th} \text{, } p_1(t)\in \left[p_{1,min} \text{, } p_{1,max}\right] \}$ such that $S_u \cap R=\emptyset$ and $0 \leq p_{1,min} \textless p_{1,max} \leq p_{end}$. We select the initial set $I=\{x \in \mathbb{R}^4|p_1=0\text{, } v_1=25\text{, } p_2\in \left[-24.5 \text{, } -23.5\right]\text{, } v_2\in \left[24.5 \text{, } 25.5\right] \}$ such that vehicle $C1$ may need to choose different behaviors according to the system state. 

\begin{figure}[htbp]
\centering
\includegraphics[scale=0.48]{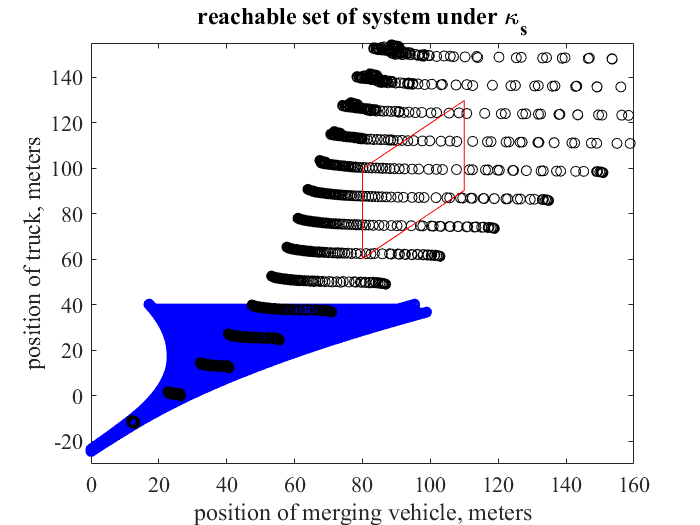}
\caption{Reachable set and sampled trajectories for the highway merging system under a single neural network planner $\kappa_s$. Due to the increasing approximation error, reachability analysis is interrupted and the blue region only shows an overapproximated subset of the reachable set where the position of truck $p_2$ is within 40 meters. The black region is 100 sampled trajectories from the same initial set $I=\{x \in \mathbb{R}^4|p_1=0\text{, } v_1=25\text{, } p_2\in \left[-24.5 \text{, } -23.5\right]\text{, } v_2\in \left[24.5 \text{, } 25.5\right] \}$. An example of unsafe set is $S_u=\{x \in \mathbb{R}^4|| p_1(t) - p_2(t)| \leq 19.75 \text{, } p_1(t)\in \left[80 \text{, } 110\right] \}$, and it is marked with a red parallelogram.}
\label{fig:sing_merging}
\end{figure}

\begin{figure}[tbp]
\centering
\includegraphics[scale=0.48]{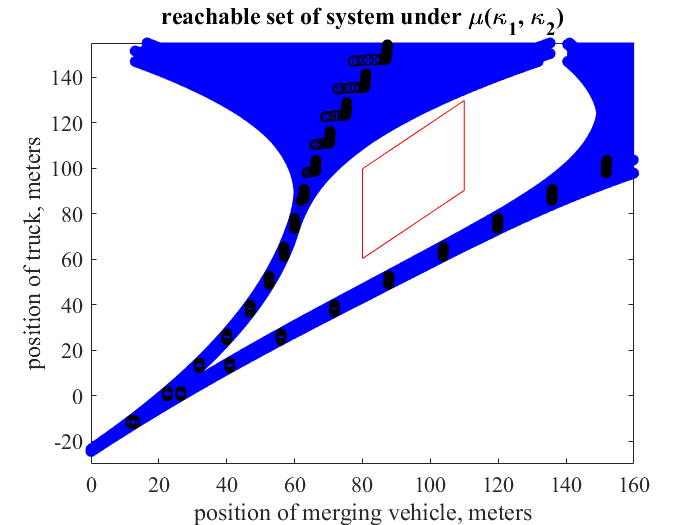}
\caption{Reachable set and sampled trajectories for the highway merging system under a hierarchical neural network planner $\mu(\kappa_1 \text{, } \kappa_2)$. The blue region is the overapproximated reachable set and the black region is 100 sampled trajectories from the same initial set $I=\{x \in \mathbb{R}^4|p_1=0\text{, } v_1=25\text{, } p_2\in \left[-24.5 \text{, } -23.5\right]\text{, } v_2\in \left[24.5 \text{, } 25.5\right] \}$. 
%States in the initial set $I$ is assessed by behavior planner $\mu$ and in this example $S_{ctrl}$ includes two behaviors of vehicle $C1$ (proceed and yield). At following time steps, vehicle $C1$ will not adjust its behavior and the reachable set of different behaviors are computed independently. 
In this case, we can find an unsafe set $S_u=\{x \in \mathbb{R}^4|| p_1(t) - p_2(t)| \leq 19.75 \text{, } p_1(t)\in \left[80 \text{, } 110\right] \}$, and it is marked with a red parallelogram. Since $S_u \cap R=\emptyset$, vehicle $C1$ can safely merge onto the highway when $p_1(t)\in \left[80 \text{, } 110\right]$.}
\label{fig:hier_merging}
\vspace{-12pt}
\end{figure}

\subsubsection{Empirical Study of Single Neural Network Planner}\label{sec:experiment single_merging}
Similarly as the unprotected left turn case study, we first consider the highway merging system under a single neural network planner $\kappa_s$, and 
Fig.~\ref{fig:sing_merging} shows the overapproximated reachable set in this case. The reachability analysis is interrupted due to increasing approximation error, and thus the blue region only shows the a subset of the reachable set where the position of truck $p_2$ is within 40 meters. This is consistent with our analysis that extremely large Lipschitz constant in this case may greatly increase the difficulty in verification. The black region is the sampled trajectories, which should be strictly covered by the reachable set (if it were computed). From this figure, we cannot find an unsafe set $S_u$ that has no overlap with the reachable set, and the system is analyzed to be unsafe under the planner $\kappa_s$.

\subsubsection{Experiment Results of Hierarchical Planner}
We then consider the highway merging system under our design of a hierarchical neural network based planner $\mu(\kappa_1 \text{, } \kappa_2)$. 
We use the same method in Section~\ref{sec:planner design} to compute the overapproximation of the reachable set in this case, as shown in Fig.~\ref{fig:hier_merging}. We assume that the behavior planner $\mu$ is triggered only once at the beginning. The left and right branch correspond to yield and proceed behavior of vehicle $C1$, respectively. We can find an unsafe region marked by a red parallelogram, $S_u=\{x \in \mathbb{R}^4|| p_1(t) - p_2(t)| \leq 19.75 \text{, } p_1(t)\in \left[80 \text{, } 110\right] \}$. Since this $S_u$ has no overlap with the overapproximated reachable set, vehicle $C1$ can safely merge into the highway when $p_1(t)\in \left[80 \text{, } 110\right]$. This case study once again demonstrates the verifiability and safety of our hierarchical neural network planner.

\section{Conclusions}\label{sec:conclusion}
We presented a hierarchical neural network based planner design based on the underlying physical scenarios of the system, and developed novel overapproximation techniques for the reachability analysis of such hierarchical design to ensure system safety. Through theoretical analysis, we showed that our hierarchical design can improve the safety and verifiability for systems that may evolve into different physical scenarios, compared with single neural network based planners. Through two case studies of unprotected left turn and highway merging in autonomous driving, we further demonstrate such advantages empirically. 

%We formally prove that our hierarchical planner has better performance than single neural network based planner regarding safety and verifiability. Case studies for unprotected left turn and highway merging demonstrate that our design significantly outperforms the baseline, which is consistent with our proof.

%\begin{acks}
%To Robert, for the bagels and explaining CMYK and color spaces.
%\end{acks}

\bibliographystyle{ACM-Reference-Format}
\bibliography{main}

%\appendix


\end{document}